\documentclass[10pt,twocolumn,letterpaper]{article}

\usepackage{iccv}
\usepackage{times}
\usepackage{epsfig}
\usepackage{graphicx}
\usepackage{amsmath}
\usepackage{amssymb}

\usepackage{color}
\usepackage{subfigure}
\usepackage[autostyle]{csquotes}
\usepackage{authblk}
\usepackage{multirow}
\usepackage{setspace}
\usepackage[pagebackref=true,breaklinks=true,letterpaper=true,colorlinks,bookmarks=false]{hyperref}

\definecolor{R}{rgb}{0.9,0.3,0.3} 
\definecolor{G}{rgb}{0.3,0.9,0.3} %
\definecolor{B}{rgb}{0.3,0.3,0.9} 

\usepackage[breaklinks=true,bookmarks=false]{hyperref}

\iccvfinalcopy 

\def\iccvPaperID{****} 

\ificcvfinal\fi

\begin{document}

\title{ILSH: The Imperial Light-Stage Head Dataset for Human Head View Synthesis}



\author[1]{Jiali Zheng}
\author[1]{Youngkyoon Jang}
\author[1]{Athanasios Papaioannou}
\author[1]{Christos Kampouris}

\author[1]{Rolandos Alexandros Potamias}
\author[1]{Foivos Paraperas Papantoniou}
\author[1]{Efstathios Galanakis}

\author[2]{Ale\v{s} Leonardis}
\author[1]{Stefanos Zafeiriou}

\affil[1]{Imperial College London, South Kensington, London SW7 2BX, UK}
\affil[2]{University of Birmingham, Edgbaston, Birmingham B15 2TT, UK}

\affil[1]{\tt\small \{ jiali.zheng18, 
youngkyoon.jang11,
a.papaioannou11, 
c.kampouris12,
r.potamias19,
f.paraperas-papantoniou22,
s.galanakis21, s.zafeiriou\}@imperial.ac.uk}
\affil[2]{\tt\small a.leonardis@cs.bham.ac.uk}


\maketitle

\ificcvfinal\thispagestyle{empty}\fi

\begin{abstract}
  This paper introduces the Imperial Light-Stage Head (ILSH) dataset, a novel light-stage-captured human head dataset designed to support view synthesis academic challenges for human heads. The ILSH dataset is intended to facilitate diverse approaches, such as scene-specific or generic neural rendering, multiple-view geometry, 3D vision, and computer graphics, to further advance the development of photo-realistic human avatars. This paper details the setup of a light-stage specifically designed to capture high-resolution ($4$K) human head images and describes the process of addressing challenges (preprocessing, ethical issues) in collecting high-quality data. In addition to the data collection, we address the split of the dataset into train, validation, and test sets. Our goal is to design and support a fair view synthesis challenge task for this novel dataset, such that a similar level of performance can be maintained and expected when using the test set, as when using the validation set. The ILSH dataset consists of 52 subjects captured using 24 cameras with all 82 lighting sources turned on, resulting in a total of 1,248 close-up head images, border masks, and camera pose pairs. 
\end{abstract}

\section{Introduction}

\begin{figure}[h]
\begin{center}
   \includegraphics[width=0.99\linewidth]{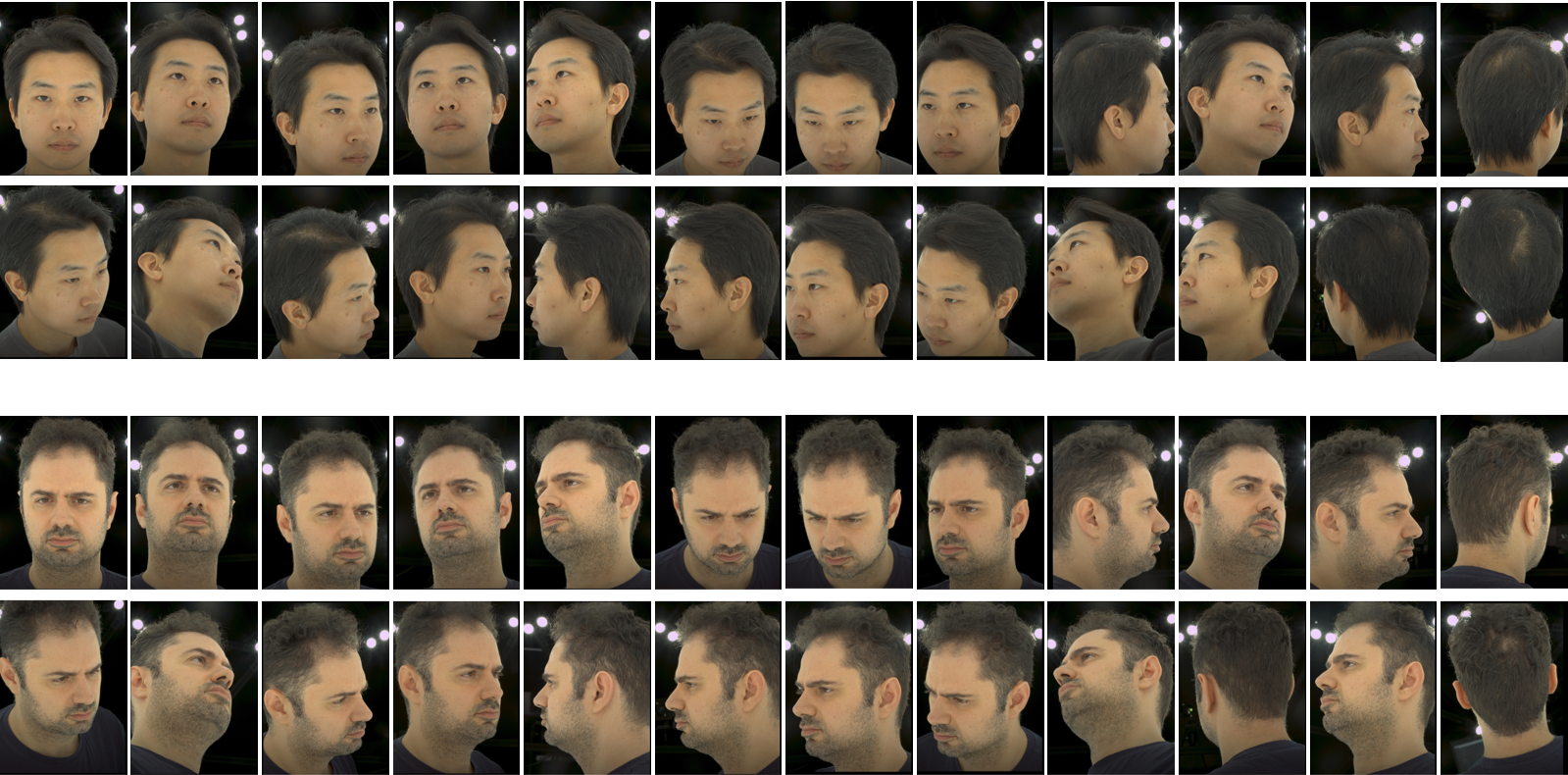}
\end{center}
   \caption{Exemplar data from the ILSH dataset.}
\label{fig:toy_example}
\end{figure}

Recent advances in novel view synthesis techniques and high-quality light-stage datasets have unlocked exciting new possibilities for photo-realistic human head avatars. In particular, the recent research boost leveraging the Neural Radiance Field (NeRF) model~\cite{mildenhall2020nerf} exhibits its potential for future synthesis avatars. Building on two decades of research efforts since the first generation of light-stage data captures~\cite{Wenger:Lightstage:TOG05, Chabert:RLHumanLocomo:SIGGRAPH06}, research communities are now focusing on finding efficient and diverse ways to solve the novel view synthesis problem while creating high-quality avatars. For example, there are various approaches being explored to tackle challenges, such as sparse view~\cite{yu2021pixelnerf}, various object volume sizes~\cite{TMPEH:CVPR:2023}, generic models~\cite{lin2022efficient}, model efficiency~\cite{mueller2022instant, SunSC22}, and bridging explicit-implicit representation~\cite{sparf2023}. However, although currently available light-stage-captured head datasets are well-designed and provide extensive data, they are limited in their ability to support potential challenges and invite diverse approaches, such as the development of generic models or the use of computer graphics or conventional 3D vision methods.

Some representative light-stage human head datasets, such as HUMBI~\cite{Yu:HUMBI:CVPR20}, Mugsy v1 and v2~\cite{Bi:AnimatableFaces:TOG21}, and NeRsemble~\cite{kirschstein2023nersemble}, provide images captured from dense camera viewpoints to support the novel view synthesis task, as shown in Table~\ref{table.datasets}. These datasets have enabled advances in generating photo-realistic human head avatars benefiting from their generous setting of densely located cameras. However, they are limited in exploring further research challenges, such as sparse viewpoints or exploring the potential of generic models, due to the small number of subject identities provided. The difficulty of inviting a large number of subjects is a naturally expected challenge when collecting real human head image datasets, mainly due to ethical issues. DigiFace-1M~\cite{bae2023digiface1m} addresses these ethical issues along with potential labeling noise and data bias, and it motivates the collection of comprehensive synthetic human head image datasets. However, using models trained on synthetic data for practical use cases will require further complicated engineering work, including an additional fine-tuning process, as real head images have different texture distributions from those found in synthetic head image datasets.

Recently, the RenderMe-360 dataset~\cite{2023renderme360} was introduced, which exhibits a large collection of diversified real human head images. The dataset appears to provide an extensive and well-balanced distribution of demographics, including age, height, weight, gender, ethnicity, as well as head accessories and hairstyles. To achieve a balanced demographic distribution in the dataset, especially for hairstyles, artificial accessories such as wigs were used, while also allowing for face tattoos and heavy makeup. These efforts also exhibit the value of collecting diversified real human head images while keeping the balanced demographics. Although a complete and balanced dataset will ultimately be beneficial, a dataset that exhibits the natural distribution of demographics can be more meaningful in attracting diverse aspects of the real challenges of novel view synthesis, instead of artificially enforcing balanced demographics.

Through this reasoning process, we collected high-quality head data (with a resolution of 3000$\times$4096) from 52 subjects using 24 cameras in our light-stage installation, as shown in Fig.~\ref{fig:ls-install}. We tried to invite a balanced number of demographic factors, but still left the natural biases, such as age and gender, that inherently arise from collecting data at an academic university location. Using this dataset, potential methods can tackle diverse challenges presented by the dataset, such as sparse viewpoints and the need for generic models, to advance photo-realistic human head avatars as a novel view synthesis challenge. In this paper, we introduce the first of its kind dataset allowing diverse approaches, called the Imperial Light-Stage Head (ILSH) dataset, which presents appropriate technical challenges while also providing room to attract diverse approaches, including generic and scene-specific models from both NeRF and non-NeRF models.

The main contributions of our work are three-fold:

\noindent {\textbf{1) Well-designed light-stage:}} We designed and installed our light-stage to collect a challenging but approachable dataset. In addition, we conducted extensive preliminary testing on data split for train, validation and test sets to guarantee a fair comparison using our dataset. 

\noindent {\textbf{2) High-fidelity dataset for diverse scenarios:}} We intentionally invited an adequate number of subjects for the dataset collection to support not only scene-specific methods but also generic models, while naturally exhibiting the challenges of sparse viewpoints, which is one of the main challenging issues of novel view synthesis.

\noindent {\textbf{3) Easy-to-use dataset:}} By providing a starter kit that includes supporting scripts for tasks such as re-structuring the downloaded dataset, loading data, and evaluating results, we help researchers lower the barrier to using our datasets to develop their methods.




For the rest of this paper, we survey related datasets (Sec.~\ref{sec.relatedwork}), introduce the preparation steps for data collection (Sec.~\ref{sec.data_col_prep}), and present the ILSH dataset (Sec.~\ref{sec.ILSH}). Then, we conclude this paper by discussing ethical considerations (Sec.~\ref{ethical considerations}). 




\begin{table*}
\begin{center}
\scalebox{0.85}{
\begin{tabular}{|l|c|c|c|c|c|c|c|c|c|}
\hline
Method & 
        Relightable & Year & Camera Pose & Video & Public & 
        \# View & \# Identity & \# Expression & Resolution  \\
\hline\hline
FaceWarehouse~\cite{Cao:FaceWarehouse:TVCG14} & 
        N & 2014 & N & N & Y & 
        1 & 150 & 20 & 640$\times$480 \\
HUMBI~\cite{Yu:HUMBI:CVPR20} & 
        N & 2020 & Y & N & Y & 
        32 & 772 & 20 & 200$\times$250 \\
Mugsy v1~\cite{wuu2022multiface} & 
        Y & 2021 & Y & Y & Y & 
        40 & 13 & 65 & 2048$\times$1334 \\
Mugsy v2~\cite{wuu2022multiface} & 
        Y & 2021 & Y & Y & Y & 
        150 & 13 & 118 & 2048$\times$1334 \\
NeRSemble~\cite{kirschstein2023nersemble} & 
        N & 2023 & Y & Y & N & 
        16 & 222 & 9 + 4 & 3028$\times$2200  \\
RenderMe-360~\cite{2023renderme360} & 
        N & 2023 & Y & Y & N & 
        60 & 500 & 6000 & 2448$\times$2048 \\  
ILSH (ours) & 
        N & 2023 & Y & N & Y & 
        24 & 52 & 1 & 3000$\times$4096 \\
            
\hline
\end{tabular}
}
\end{center}
\caption{Light-stage datasets for human head.}
\label{table.datasets}
\end{table*}

\section{Related Datasets}
\label{sec.relatedwork}

\textbf{Light-stage human head and body datasets.} Efforts to capture high-quality human data have been aligned with the history of light-stage data capture~\cite{Wenger:Lightstage:TOG05, Chabert:RLHumanLocomo:SIGGRAPH06}. Light-stages are typically designed to capture synchronized illumination conditions and images for relighting humans. However, they have recently been used to capture ground truth data for novel view synthesis research tasks, since they provide high-quality camera calibration and controlled data capture. As traditional light-stage data capture focused on capturing the human head~\cite{Wenger:Lightstage:TOG05} and body data~\cite{Chabert:RLHumanLocomo:SIGGRAPH06} separately, the current threads of human datasets continue to advance in the mainstream of collecting human head~\cite{Cao:FaceWarehouse:TVCG14, Yu:HUMBI:CVPR20, Sun:LSSuperRes:TOG20, Bi:AnimatableFaces:TOG21, wuu2022multiface, Wang:MoRF:SIGGRAPH22, 2023renderme360} and body~\cite{Guo:Relightables:TOG19, Remelli:VolumAvatar:SIGGRAPH22, isik2023humanrf, zhou2023relightable, Cheng_iccv2023dnarendering} images and videos. In this section, we review related datasets that focus on collecting real human head data.


\textbf{Resolutions and viewpoints.} 
FaceWarehouse~\cite{Cao:FaceWarehouse:TVCG14} is a database of 3D human faces that provides RGBD face data with a resolution of 640$\times$480 and estimated 3D geometry. Although it includes 150 subjects, only one view is available for each subject, which is insufficient for studying human head neural rendering from multi-view images. HUMBI~\cite{Yu:HUMBI:CVPR20}, a large-scale multi-view human dataset aimed at facilitating high-resolution human body appearance learning, used a 107 dense camera array to capture the gaze, face, hand, body, and garment of 772 subjects. The resolution of the captured images for the entire body is 1920$\times$1080, while the face images are cropped to a resolution of 200$\times$250. This face data does not provide enough detail of the human head, making high-fidelity human head neural rendering challenging.

Mugsy v1 and v2 (Multiface~\cite{wuu2022multiface}) are notable human head datasets that feature head images with 2048$\times$1334 resolution of 13 individuals, with the goal of creating high-quality human head avatars using 150 camera viewpoints for 118 diverse expressions. However, with recent advances in neural rendering for novel view synthesis, datasets that present more challenges by using relatively sparse viewpoints and a more diverse set of individuals are becoming increasingly important factors to consider in order to invite novel approaches tackling practical issues in the field of human head neural rendering research. NeRSemble dataset~\cite{kirschstein2023nersemble} appears to introduce 16 sparse camera viewpoints with a large number (222) of individuals to support multi-view radiance field reconstruction. They provide high-fidelity images with 3028$\times$2200 resolution. However, the actual distances between these 16 cameras are as closely dense as those in the Mugsy v2 dataset~\cite{wuu2022multiface}, as the cameras are located within the frontal area of the face. 


\textbf{Extensive collection.} As previously discussed, RenderMe-360~\cite{2023renderme360} has made significant efforts to collect a near-completely balanced demographic for various aspects, such as age, height-weight, gender, and ethnicity, for a large number (500) of individuals. The dataset also allows for some artificial accessories and face tattoos, which may partially obscure the natural texture of human heads. Although the dataset is comprehensive and near-complete in terms of human head diversity, we decided to allow for some naturally occurring biases while doing our best to invite voluntary subjects to participate in the ILSH data collection, as human head datasets must be collected and used ethically. Additionally, these natural biases present in the dataset can attract further research questions, while inviting our originally planned potential approaches, such as handling sparse viewpoints and using generic models with NeRF or non-NeRF methods. RenderMe-360 has not yet been fully released and is planned to be released at the end of 2023.

\section{Data Collection}
\label{sec.data_col_prep}

\begin{figure*}[t]
\begin{center}

   \includegraphics[width=0.99\linewidth]{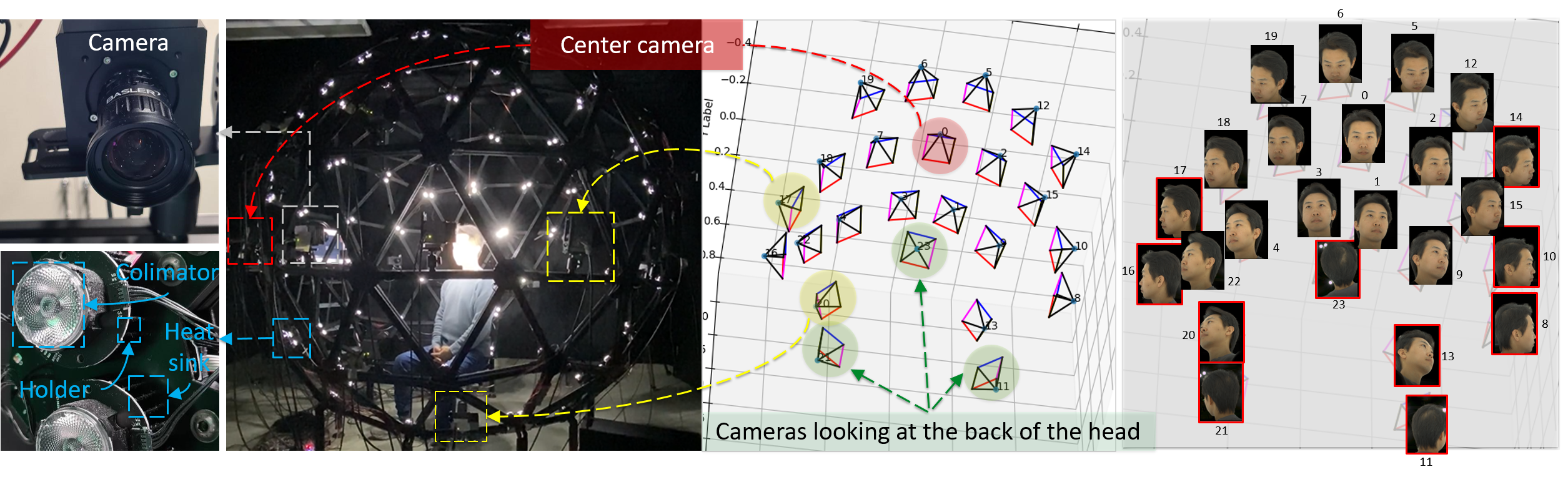}
\end{center}
   \caption{Example of a light-stage capture environment and multiple viewpoints of a subject in the ILSH dataset. The red rectangular outlines on multiple view images, as shown in the right-most image, represent the views that were excluded from the validation and test views selection, based on the distance between adjacent views, as shown in Fig.~\ref{fig:dist_btw_cams}. The remaining viewpoints are being tested across subjects as both validation and test sets.}
\label{fig:ls-install}
\end{figure*}

\noindent{\textbf{Overall process. }}
We collected high-quality images of 52 human heads over four different sessions, each corresponding to a different date (specifically, February 3rd, 10th, 16th, and 17th). Our data collection process consisted of the following steps: 1) calibration checkerboard capture, 2) color chart capture, and 3) subject capture. The first and the second steps were performed before each recording session using a 10$\times$10 20 mm checkerboard and a color chart~\cite{spyder_checkr}, respectively. This was done to ensure that the most accurate possible images and camera extrinsic parameters were provided, as well as consistent colors across images captured from individual cameras. These factors are important for the novel view synthesis task and to avoid issues such as camera lens looseness or physical displacement that might occur over time. We ultimately computed and provided four different sets of camera poses for the four different sets of subjects, aligned with the capture dates corresponding to each session.

After completing the first two steps in each session, we invited participants to have their head images captured using our light-stage (See details in Sec.~\ref{subsec.Light-Stage Setup and Installation}). Once participants were informed about the research and agreed to participate by signing the consent form (See details in Sec.~\ref{ethical considerations}), we proceeded with capturing high-quality human head images (the third step). Individual subjects were instructed to sit on a chair located in the middle of the light-stage, with their face centered within the half-sphere where 21 cameras were located, while 3 cameras captured the back of their head. The instructor in the room advised on how to best align the subject’s head by monitoring three external displays outside the light-stage that previewed images captured by orthogonally located cameras, ensuring that the subject’s head was centered before capturing the images. Once the instructor confirmed the head location, the ambient lighting in the room is turned off to avoid external light sources, and image capturing proceeds. 

\subsection{Multi-view Image Capture System}
\label{subsec.Light-Stage Setup and Installation}
To collect the ILSH dataset, we used a light-stage, as shown in Fig.~\ref{fig:ls-install}. The frame of the light-stage is based on a 3-frequency geodesic dome with a diameter of 2.5 m, and its bottom part is cropped to accommodate a chair. When the subject sits on the chair, their head is positioned in the center of the dome at a height of 1.25 m. The capture system is located in a room with matte black walls to minimize reflections. 

The current version of the system consists of 82 light sources with high-power RGBW LEDs (OSRAM OSTAR) and narrow-angle lenses to concentrate the light on the subject’s head. For data capture, we used only the white LEDs of the light sources with fixed intensity to create uniform illumination.

The light-stage is equipped with 24 machine vision cameras (Balser boA4112-68cc) with a resolution of 4096$\times$3000 and a global shutter. Among them, 21 cameras are located on the front hemisphere, looking at the subject’s face, while the remaining three are looking at the back of the head, as discussed above. The cameras are synchronized with a bespoke hardware trigger system that ensures all images are captured simultaneously. To capture high-quality data, we used 50 mm lenses for tight framing and saved all images as RAW files.

\subsection{Camera Calibration and Color Correction}
\label{subsec.before_actual_dataset_capture}

\noindent{\textbf{Camera calibration.}}
For the checkerboards, we captured 10 mostly frontal images per camera, with slight 3D rotation, to detect a sufficient number of stable corner points from a single view. As previously mentioned, all cameras were synchronized during the capture. In total, we captured 5,760 checkerboard images (10$\times$24 directions$\times$24 cameras), but only used those where the checkerboard was accurately and stably detected.

For the color chart board capture process, we followed the same procedure as for capturing checkerboards. However, we only captured 2 frontal color chart images per camera, resulting in a total of 48 images (2$\times$24). This was done to correct the colors of individual cameras using a standard color chart board as a reference, ensuring consistency in the colors captured by separate cameras.

For camera calibration, we used a multi-camera calibration method~\cite{wang2022accurate} based on checkboard patterns. In the process, we manually selected an image, which serves as a reference coordinate system, among many checkerboard images captured by the first camera (View-ID: 0 in Fig.~\ref{fig:ls-install}) located in the center of the first hemisphere. The first camera subsequently captures the frontal face images of a subject’s head for our dataset. Given the intrinsics and extrinsics from the calibration process, we undistorted the images and finally released the dataset with border masks, which were produced during the undistortion process to mask black empty areas in the undistorted images. To support neural rendering-based methods (e.g., vanilla NeRF~\cite{mildenhall2020nerf}), we provide camera pose information in commonly used formats such as LLFF and Blender, containing camera intrinsics and extrinsics, near and far values, image resolution, and focal length.

\noindent{\textbf{Color correction.}}
In addition to camera calibration, we corrected the color to achieve the most realistic (natural) color using a standard color chart template. We estimated the color transform coefficients between the reference color chart and the input color from light-stage cameras using Cheung’s method~\cite{Cheung_CC_CT04,mansencal_thomas_2022_7367239}.
The camera and illumination emitter (LED) settings were set to ISO 200, an exposure time of 20 ms (shutter speed 50 FPS), and 4\% individual light emitter strength. These parameters were fixed for capturing the dataset for all subjects.

\section{ILSH: Imperial Light-Stage Head Dataset}
\label{sec.ILSH}

In this section, we first describe how we prepared additional image border masks and performed the undistortion process. We then provide an overview of our final ILSH dataset. We also discuss how we split the data into train, validation, and test sets to ensure fairness in academic Challenges. This is particularly important in an open research competition where the test set is completely hidden, allowing participants to have a fair chance to achieve the expected test results of their methods, which were developed and tested using the validation set. Finally, we describe the demographics of our dataset.

\subsection{Preprocessing}
\label{subsec. prep_challenge}

To prepare our dataset for release, we followed four preprocessing steps for the entire dataset: 1) camera calibration, 2) color correction, 3) dataset anonymization, and 4) data split into train, validation, and test sets. After capturing the checkerboard and color chart images, as discussed in Sec.~\ref{subsec.before_actual_dataset_capture}, we conducted the first camera calibration and the second color correction steps using the captured images. Then, after camera calibration and color correction, we undistorted the final images using the distortion coefficients optimized in step 1. To simplify dataset usage, we unified the intrinsic parameters among all cameras (to the average camera matrix) and remapped the color images to correspond to the new intrinsics. This undistortion and remapping shifted pixels, often resulting in a band of empty pixels at some image borders. We therefore provide masks, referred to as border or undistortion masks, with a value of one for valid pixels and zero for empty pixels. To anonymize the data, we renamed all images using the format: $\{:03d\}\_00\_\{:02d\}.format(subject\_id, view\_id)$, resulting in names such as $000\_00\_00.png$ for subject 0 captured by camera 0, which is the first camera, as an example. 

The data is structured as follows. Each subject-specific directory (e.g., $000\_00$) contains an \emph{images} and a \emph{masks} folder, holding the undistorted, renamed head images and the border masks produced during the undistortion process, respectively. The masks follow the same naming format as the images but with a .npy file extension. Each subject folder also includes camera intrinsics and extrinsics. 
As mentioned in Sec.~\ref{subsec.before_actual_dataset_capture} we provide the camera information in two commonly used camera conventions: LLFF ($poses\_bounds\_\{train, val, test\}.npy$) and Blender ($transforms\_\{train, val, test\}.json$). 

\begin{figure}[t]
\begin{center}
   \includegraphics[width=1\linewidth]{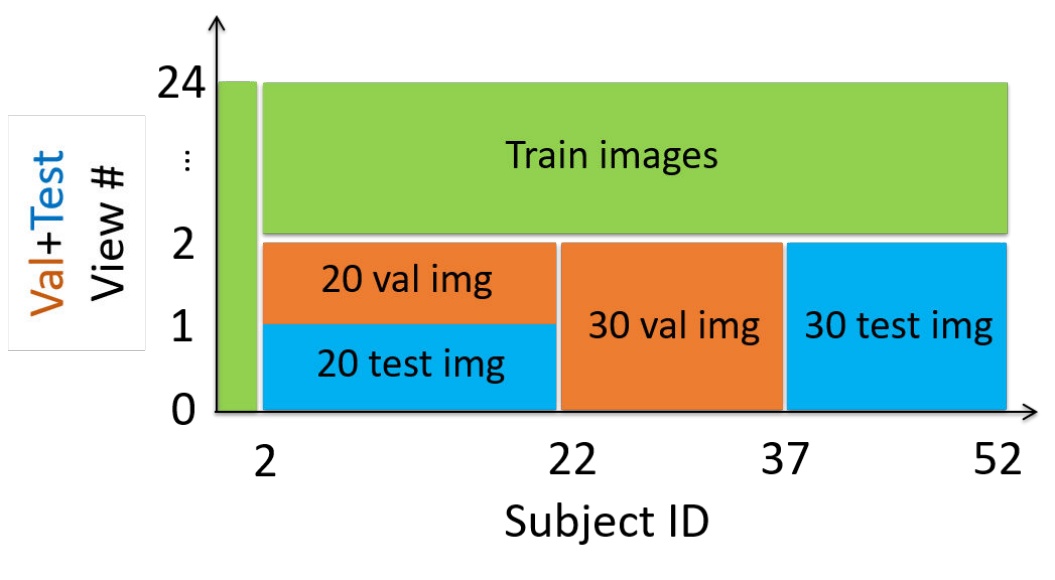}
\end{center}
   \caption{Demographics of the data split into train, validation, and test sets.}
\label{fig:data_split}
\end{figure}

\begin{figure}[t]
\begin{center}
   \includegraphics[width=1\linewidth]{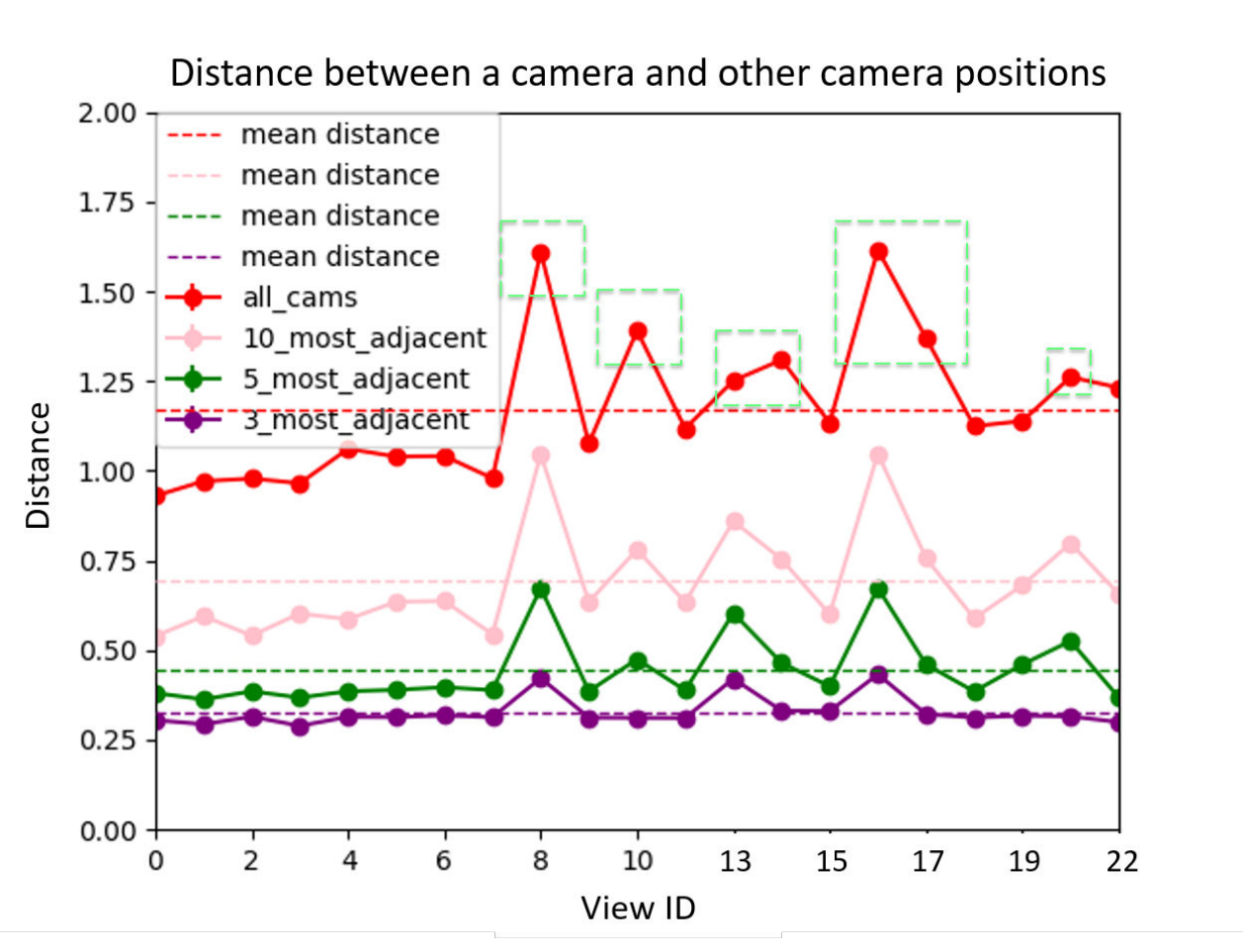}
\end{center}
   \caption{Distance between a camera and other camera positions, excluding the 3 backside views with IDs are 11, 21, and 23.}
\label{fig:dist_btw_cams}
\end{figure}

\subsection{The ILSH Dataset}
\label{sec.ilsh_dataset}
\noindent{\textbf{Overview of the dataset. }}
After going through all the preprocessing steps, we collected and completed the ILSH dataset. The ILSH dataset includes \textbf{52 subjects}, each represented in a separate folder. A total of \textbf{24 cameras} were utilized to capture images of the individual subjects, so each folder has 24 head images captured from different viewpoints. In total, the dataset contains \textbf{1,248 4K resolution (width: 3000 and height: 4096) images} of the 52 subjects, captured using the 24 cameras. Due to calibration-based image undistortion preprocessing applied to the dataset, some image pixels along the borders will remain empty (See details in Sec.~\ref{subsec. prep_challenge}). Thus, we provide \textbf{1,248 binary border masks} resulting from this calibration, with a value of 1 for non-empty pixels and 0 for empty pixels, along with the head images. 

In addition to the head images and their corresponding border masks, we also provided \textbf{1,248 Camera Poses} in multiple formats. Each subject folder contains two formats of camera pose files: one in a Blender-compatible data loader format (transforms.json) and one in an LLFF-compatible data loader format (pose\_bounds.npy). Each file contains 24 camera poses that are dependent on the capture date, along with additional information such as focal length and image resolution. In total, there are 52 pairs of these files in their respective subject folders.


\begin{figure*}[t]
\begin{center}
   \includegraphics[width=1\linewidth]{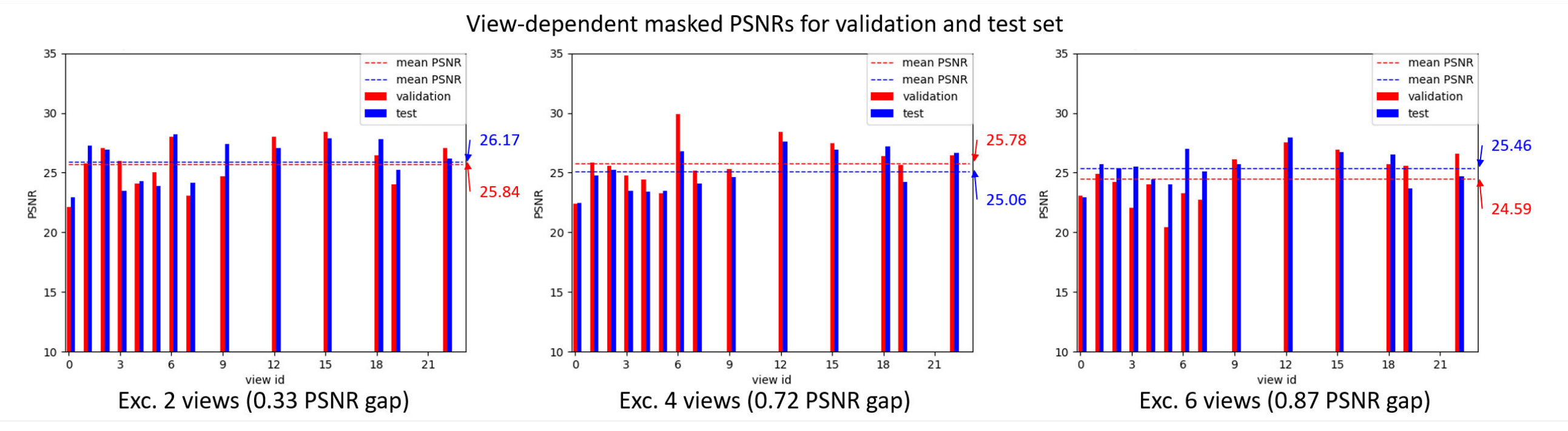}
\end{center}
   \caption{Effect of changing the number of excluded views used in validation and test sets (left: 2, middle: 4, and right: 6 views). Increasing the number of views excluded from training results in a larger gap in performance between validation and test sets, leading to a failure to maintain performance consistency.}
\label{fig:effect_of_No_exc_views}
\end{figure*}

\noindent{\textbf{Data split for train, validation, and test sets. }} 
We split the dataset into three subsets: train, validation, and test sets, as shown in Fig.~\ref{fig:data_split}. This data split is designed to ensure fair comparison for a novel view synthesis challenge task given 52 subjects. 
We made every effort to maintain consistent performance between the validation and test sets when selecting views to assign to each subset. 
To achieve this, we decided to assign a minimal number (two) of views out of the 24 available views for each subject in both the validation and test sets. This guarantees a consistent number of training views for all subjects.

In addition to maintaining the same number of training samples within each subject, we were careful to assign a fairly equal number of validation and test views across subjects. This ensured that the training, validation, and test sets had an equal amount of data samples for each viewpoint. To make the assigning view process even fairer and to preserve similar performance across the validation and test sets, we also had to avoid exceptionally distant views (which are sparse viewpoints) from the pool of validation and test sets. 
We believed that one way to further ensure consistent performance between the validation and test sets was to remove any additional factors that could introduce random difficulties, such as varied view sparsity among the validation and test sets. To do this, we calculated the mean distance between 3, 5, 10, and all adjacent camera locations from a given camera location, as shown in Fig.~\ref{fig:dist_btw_cams}. Then, we discarded 10 views in descending order of their mean distances. These views are depicted as red rectangular outlines on multiple view images in Fig.~\ref{fig:ls-install}.

Upon examining the calculated mean distances, we confirmed that the discarded distant views correlated with a low density of surrounding cameras. These views were mainly from the back and profile side views, where one side of a camera is not fully surrounded by other cameras. As a result, all remaining validation and test views are now fairly surrounded by other cameras, which will aid in training for a novel view synthesis task. Although we discarded exceptionally distant camera views for the validation and test sets, the remaining 14 candidate views still vary in difficulty according to the variance in their sparsity.

\noindent{\textbf{Minimal number of views excluded during training. }} Given this filtered pool of 14 candidate views, we tested how many excluded views would produce more challenges while still preserving the consistency of performance between the validation and test results, as shown in Fig.~\ref{fig:effect_of_No_exc_views}. We found that when we excluded the minimal number of views (two views) during training, the consistency of performance between the validation and test results was better preserved compared to when we excluded 4 or 6 views, while still being challenging enough. Otherwise, when the number of excluded views during training increases, the performance gap appears to immediately become larger. Thus, we decided to assign a minimal number of views to be selected for the validation and test sets. As a baseline method, we used TensoRF~\cite{Chen2022ECCV}. As shown in Fig.~\ref{fig:effect_of_No_exc_views}, when excluding 2, 4, and 6 views for the validation and test sets in the training set, the average score gaps between the validation and test sets were 0.33, 0.72, and 0.87, respectively. Upon closer examination of the graph, we confirmed that the score gaps for each view, such as camera view 6, worsened as the number of excluded views increased.

\noindent{\textbf{Three subject categories. }} In our dataset, there are three categories of subjects: 1) those who have 1 validation and 1 test view, 2) those who have only 2 validation views, and 3) those who have only 2 test views, as shown in Fig.~\ref{fig:data_split}. The first category enables researchers to develop algorithms and train models that have been validated using the validation set and can be reused for the test set to maintain their performance on the same subjects. The second and third categories are useful for confirming the model’s generalizability when tested across subjects with different appearances. Having these three types of subject categories, while assigning validation and test views across subjects, helps us to better evaluate the generalized performance of models, which is independent of both view sparsity and subject appearance.

Due to the lengthy training time typically associated with conventional neural rendering models, we released two toy examples with full camera poses for simpler testing.




\noindent{\textbf{Algorithm to assign a balanced number of validation and test views for each subject. }} 
We followed these key steps to select the most distant two views out of 14 available views per subject across 50 subjects: 
\noindent{\textbf{1.}} For the first subject, randomly choose one view from the 14 available views.
\noindent{\textbf{2.}} Make a list of the remaining 23 views (as pairs containing the chosen view) in order of their distance from the chosen initial view.
\noindent{\textbf{3.}} Examine the list from the top to see if any combination of two views has not been previously selected as a combination for validation and test views.
\noindent{\textbf{4.}} If a combination has not been selected before, add it to the list of selected view combinations and remove it from the available list of views.
\noindent{\textbf{5.}} Increase the subject ID and repeat the process for each subject until all subjects have been processed.
\noindent{\textbf{6.}} If the available views run out while processing, refill the available views with the initial 14 views and continue processing.

If the last pair of views, which is the only choice among the unselected available views, already exists in the selected combinations, we avoid this exceptional case by reversing the order of the camera combination. As we assign views for the validation and test sets in order for the same subject, this permutation process is sufficient to avoid the exception case that may occur only once or twice. After assigning balanced validation and test views across subjects, we have 50 sets of camera view tuples. We then split these into three subcategories, as previously discussed, while being careful to avoid duplicated tuples in the same categories. As a result, we have a balanced view count for each camera view for validation and test sets.

\subsection{The ILSH Dataset Demographics}
In preparing the data collection, we aimed for diversity in subject demographics, including age, gender, and ethnicity. However, due to the nature of the high-resolution human head data collected from multiple viewpoints and the voluntary nature of participation in the study, we were unable to achieve completely balanced demographics. We made every effort to invite a diverse group of participants by publicizing the study to different age and gender groups. However, we still allowed for the natural biases that may arise from the voluntary participation in the dataset collection process. As a result, we compiled the following statistics on the demographics represented in the dataset.

\begin{figure}
\begin{center}
   \includegraphics[width=1\linewidth]{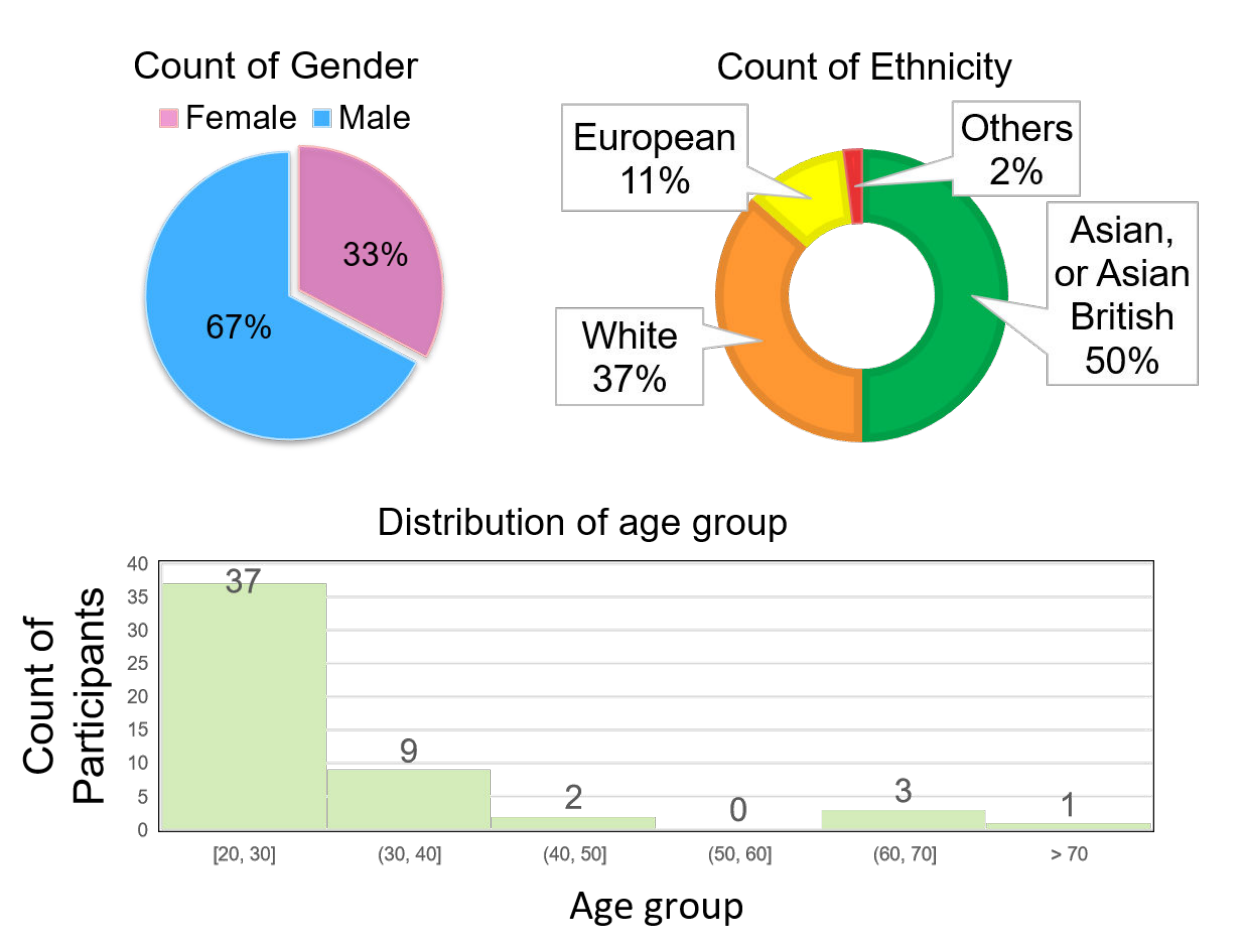}
\end{center}
   \caption{Demographics of subjects in the ILSH dataset.}
\label{fig:stats_age}
\end{figure}

Our dataset includes 52 subjects.
Within the age groups of 20s, 30s, 40s, 60s, and 70s, we have 37, 9, 2, 3, and 1 subjects, respectively. 67\% of participants are male and 33\% are female. Half (50\%) of the subjects are Asian or Asian British, 37\% are White, 11\% are European, and the remaining 2\% marked themselves as Others or chose not to reveal their ethnicity (Fig.~\ref{fig:stats_age}).

\section{Ethical Considerations}

\label{ethical considerations}

\textbf{For subjects participating in dataset collection.}
We have taken care to consider ethical considerations throughout the data collection process. When participants applied to take part in the data collection, the organizers shared an adult information sheet and a consent form with them before they arrived. The adult information sheet contains information about the experiment (data collection), including its purpose, what will happen, why we are doing it, eligibility, the option to change their mind and stop participating, potential results of the research, legal basis, what information we collect, participants’ rights to request the removal of their data at any time, information transfers, an explanation of what to do if something goes wrong, rewards, the reviewer of this study, funding body, and organizer information. The consent form contains checklists asking for approval of reading the adult sheet, having opportunities for questions and answers, its voluntary nature, human head image storage, its potential use for developing novel algorithms and training models, analysis by Imperial College, anonymization of personal data, and consent to take part in the study by providing their signature. Finally, we recruited 52 subjects for the data collection following the guidelines of the Imperial College London Ethics Committee, ensuring that all necessary precautions were taken to avoid any potential ethical mistakes in the process of hiring, collection, distribution, and maintenance.


\textbf{For the dataset user.}
In this section, we describe how we release the dataset in an ethical manner. The Imperial College London Ethics Committee and organizing team are diligently involved in preventing any potential harm, as this dataset contains personal facial information that must be used ethically. As described above, The Ethics Committee first approves the collection of this human head dataset and closely advises us to avoid any potential harm and guarantee the subjects’ (dataset participants’) right to stop participating. In addition, it is agreed that the dataset would only be shared within officially guaranteed research communities for research purposes only.

To follow the guideline of sharing within guaranteed research institutes, we do not share the dataset directly. Instead, we first require dataset users to agree to the End User License Agreement (EULA) and disclose their identity so that we can track who is using the dataset. 


\section{Conclusion and Future Work}

In this paper, we introduced the Imperial Light-Stage Head (ILSH) dataset, which is a valuable resource for advancing the development of photo-realistic human head avatars. The dataset, which was specifically designed to support view synthesis challenges for Human Heads, provides high-quality data captured using our custom light-stage setup. The ILSH dataset was also designed to facilitate diverse approaches, including scene-specific or generic neural rendering, 3D vision (e.g., multiple-view geometry), and computer graphics. The design of a fair view synthesis challenge task with a fair data split 
further enhances the value of this novel dataset. We expect that the ILSH dataset will inspire further research and innovation in the field of photo-realistic human head avatar generation.

{\small
\bibliographystyle{ieee_fullname}
\bibliography{egbib}
}

\clearpage
\appendix

\end{document}